\documentclass[10pt,twocolumn,letterpaper]{article}

\usepackage{cvpr}              

\usepackage{multirow} 
\usepackage[dvipsnames,table,xcdraw]{xcolor}

\definecolor{cvprblue}{rgb}{0.21,0.49,0.74}
\usepackage[pagebackref,breaklinks,colorlinks,allcolors=cvprblue]{hyperref}

\def\paperID{****} 
\def\confName{CVPR}
\def\confYear{2026}

\newcommand{\blfootnote}[1]{%
	\begingroup
	\renewcommand\thefootnote{}
	\footnotetext{#1}%
	\endgroup
}

\title{Breaking Down Monocular Ambiguity: Exploiting Temporal Evolution \\ for 3D Lane Detection}

\author{
Huan Zheng$^{*}$,
Wencheng Han$^{*}$,
Tianyi Yan, 
Cheng-zhong Xu,
Jianbing Shen$^{\dagger}$\\
SKL-IOTSC, CIS, University of Macau\\
}

\begin{document}
\maketitle

\begin{abstract}
	Monocular 3D lane detection aims to estimate the 3D position of lanes from frontal-view (FV) images. 
	However, existing methods are fundamentally constrained by the inherent ambiguity of single-frame input, which leads to inaccurate geometric predictions and poor lane integrity, especially for distant lanes.
	To overcome this, we propose to unlock the rich information embedded in the temporal evolution of the scene as the vehicle moves. 
	Our proposed Geometry-aware Temporal Aggregation Network (GTA-Net) systematically leverages the temporal information from complementary perspectives.
	First, Temporal Geometry Enhancement Module (TGEM) learns geometric consistency across consecutive frames, effectively recovering depth information from motion to build a reliable 3D scene representation.
	Second, to enhance lane integrity, Temporal Instance-aware Query Generation (TIQG) module aggregates instance cues from past and present frames. 
	Crucially, for lanes that are ambiguous in the current view, TIQG innovatively synthesizes a pseudo future perspective to generate queries that reveal lanes which would otherwise be missed.
	The experiments demonstrate that GTA-Net achieves new SoTA results, significantly outperforming existing monocular 3D lane detection solutions.
\end{abstract}

\blfootnote{$*$Equal contribution. $\dagger$Corresponding author: \textit{Jianbing Shen}.}

\begin{figure}[t]
	\centering
	\includegraphics[width=\linewidth]{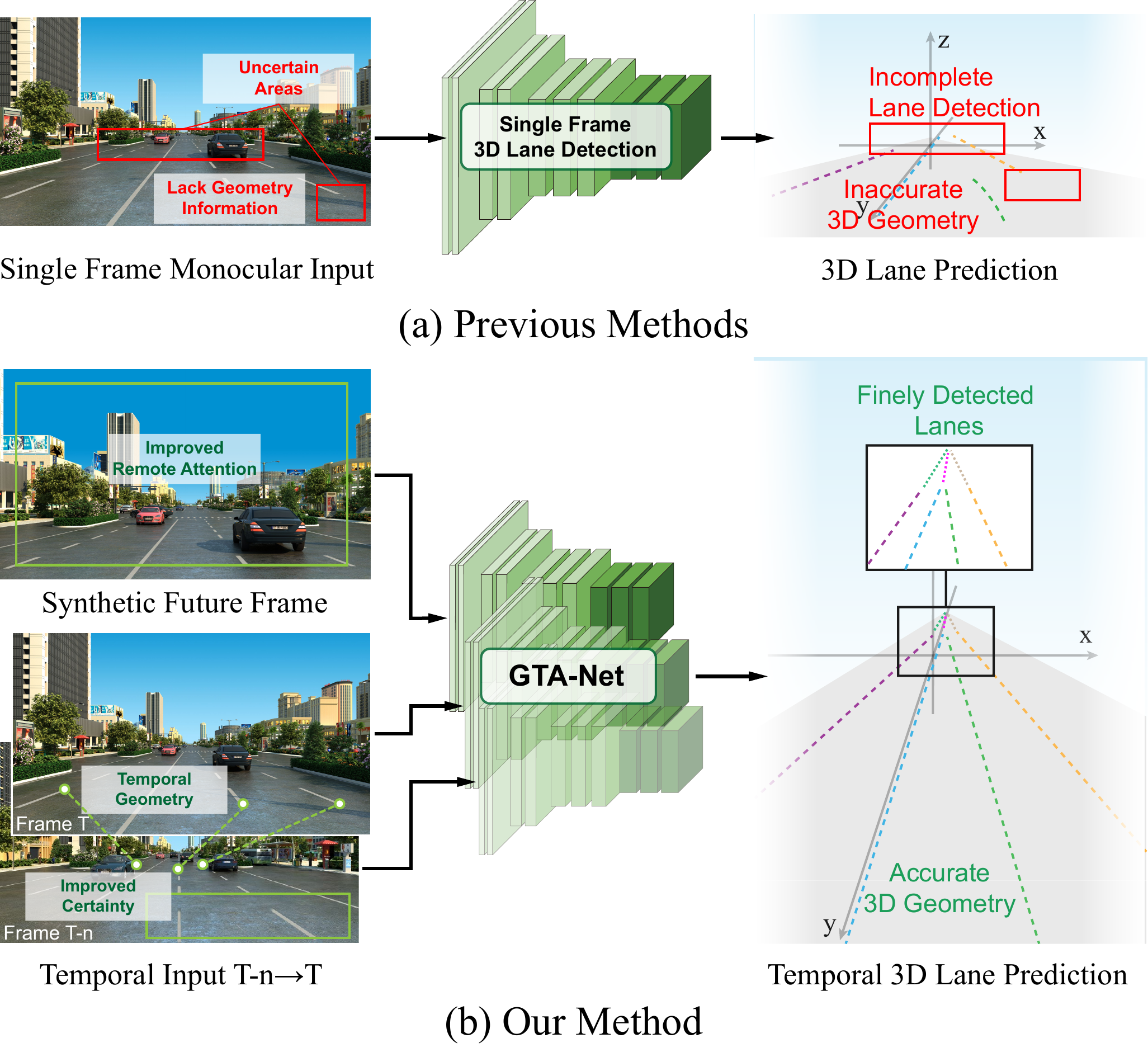} 
	\vspace{-7mm}
	\caption{\textbf{Comparison with Monocular 3D Lane Detection Approaches.} (a) Existing methods \cite{huang2023anchor3dlane, bai2023curveformer, guo2020gen, chen2022persformer, garnett20193d} that rely on a single frame as input are hindered by inaccuracies in 3D geometry perception and difficulties in maintaining lane integrity. (b) In contrast, our approach, which incorporates multiple frames along with synthetic future frames, facilitates improved 3D geometric perception and enables integral lane detection within the scene.}
	\vspace{-5mm}
	\label{fig:intro}  
\end{figure}

\section{Introduction}

Monocular 3D lane detection, the task of estimating the precise 3D coordinates of lane lines from a single camera, is a cornerstone of modern autonomous driving systems~\cite{ma2024monocular}. Accurate 3D lane perception is crucial for safe lane keeping~\cite{wang2005lane}, as well as for downstream tasks like high-definition map construction~\cite{li2022hdmapnet, liu2023vectormapnet} and trajectory planning~\cite{williams2022trajectory}. While cost-effective and rich in texture, monocular vision presents a formidable challenge: how to recover 3D structure from a 2D image.

Existing methods attempt to bridge this 2D-to-3D gap through various strategies, such as two-stage projection with depth estimation~\cite{yan2022once}, or by transforming features into a bird's-eye-view (BEV) representation~\cite{chen2022persformer, garnett20193d}. However, these approaches are built on fragile assumptions, like the reliance on highly accurate depth predictions or a flat-ground world, which often fail in complex real-world scenarios. More recent methods using 3D anchors or queries~\cite{huang2023anchor3dlane, luo2023latr} have shown promise, yet they still struggle with the fundamental ambiguity inherent in a single frame. This ambiguity manifests in two critical failures, as illustrated in Figure~\ref{fig:intro}~(a): \textbf{1) inaccurate geometry}, where predicted lanes deviate from their true 3D positions, and \textbf{2) broken lane integrity}, where lanes, especially at a distance, are partially detected or missed entirely.

We argue that the key to overcoming these limitations lies not within the single frame, but in the \textbf{\textit{temporal evolution}} of the scene. As a vehicle moves, the interplay between consecutive frames contains powerful cues to resolve the aforementioned ambiguities. In this paper, we propose the Geometry-aware Temporal Aggregation Network (GTA-Net), a novel framework designed to systematically exploit this temporal information from complementary perspectives, as shown in Figure~\ref{fig:intro}~(b).

First, to address the challenge of \textit{inaccurate geometry}, we ask: \textbf{\textit{how can the past inform the present's geometry?}} The relative motion between frames directly encodes the spatial relationships among scene elements. Inspired by this, we introduce the Temporal Geometry Enhancement Module (TGEM). By explicitly modeling the geometric consistency between past and present frames, TGEM learns to perceive depth-aware features, effectively correcting the geometric representation of the scene and leading to more accurate lane positioning.

Second, to tackle the problem of \textit{broken lane integrity}, we ask: \textbf{\textit{how can temporal context help us see ambiguous lanes?}} A lane that is faint or occluded now may have been clear in the past, or will become clear in the future. We therefore propose the Temporal Instance-aware Query Generation (TIQG) module. TIQG generates lane queries by aggregating instance information from both past and present views. More importantly, it addresses the challenge of detecting distant lanes with a novel strategy. As shown in Figure~\ref{Fig:zoomin}, a distant, undetectable lane becomes visible when the viewpoint moves closer. Since we cannot access real future frames, TIQG innovatively synthesizes a pseudo-future frame by digitally zooming into the region of interest. This allows us to generate effective queries for distant regions, dramatically improving lane integrity.

By seamlessly integrating these two components, GTA-Net leverages the temporal evolution to build a comprehensive understanding of the driving scene. Our main contributions are as follows:
\begin{itemize}
    \item We propose the Geometry-aware Temporal Aggregation Network (GTA-Net), a novel architecture that systematically leverages temporal information to overcome the limitations of monocular 3D lane detection.
    \item We introduce the Temporal Geometry Enhancement Module (TGEM), which exploits geometric consistency across frames to learn depth-aware features for accurate 3D geometry perception.
    \item We present the Temporal Instance-aware Query Generation (TIQG) module, which enhances lane integrity by aggregating temporal instance cues.
    \item Extensive experiments show that GTA-Net achieves new state-of-the-art performance on the OpenLane benchmark, demonstrating its effectiveness and robustness across diverse and challenging scenarios.
\end{itemize}

\section{Related Work}

\subsection{2D Lane Detection}
2D lane detection, a critical aspect of autonomous driving, strives to precisely determine the shapes and positions of lanes within images \cite{tang2021review, xu2020curvelane}. 
With the development of deep learning \cite{kirillov2023segment, yan2025olidm, yan2025drivingsphere, yan2025rlgf}, deep neural network-based 2D lane detection has achieved remarkable strides forward \cite{jin2023recursive, qin2022ultra, wang2022keypoint, zheng2022clrnet, han2022laneformer}.
According to the formulation of 2D lane detection, existing solutions can be divided into four categories: segmentation-based, anchor-based, keypoint-based and curve-based methods \cite{huang2023anchor3dlane}.

Segmentation-based approaches \cite{ko2021key, li2021abssnet, zheng2021resa} employ segmentation tasks as intermediaries for executing lane detection. To be specific, these methods firstly complete pixel-wise classification to obtain 2D lane segmentation masks. 
Next, the segmentation masks are processed by some carefully-design strategies to construct a set of lanes.
Anchor-based approaches primarily utilize line-like anchors for regressing offsets relative to ground truth lanes \cite{jin2022eigenlanes, liu2021condlanenet, qin2020ultra, qin2022ultra, tabelini2021keep, yoo2020end}.
To enhance the efficacy of line-shape anchors, row-based anchors have been devised for capitalizing on the natural shape of lanes \cite{yoo2020end}. 
By using row-based anchors, 2D lane detection task is converted into a row-wise classification problem.
Furthermore, these row-based and column-based anchors are combined to reduce the occurrence of localization errors in side lanes \cite{qin2022ultra}.
Keypoint-based approaches offer a flexible framework for modeling lane positions \cite{ko2021key, qu2021focus, wang2022keypoint}. 
At the outset, these methodologies embark on predicting the 2D coordinates of pivotal points along the lanes \cite{ko2021key}. 
Following this initial step, a diverse array of strategies is employed to effectively group these key points into distinct lanes \cite{wang2022keypoint}. 
Curve-based approaches leverage a range of curve formulations to model lanes \cite{feng2022rethinking, liu2021end, tabelini2021polylanenet}.
Specifically, this type of method aims to estimate the parameters of curves to obtain 2D lane detection results.
BézierLaneNet \cite{feng2022rethinking} leverages the parametric Bézier curve for its computational simplicity, stability, and extensive flexibility in transformations.
Additionally, the authors introduce a novel feature flip fusion module, which takes into account the symmetrical characteristics of lanes.

\begin{figure*}
    \centering
    \includegraphics[width=\linewidth]{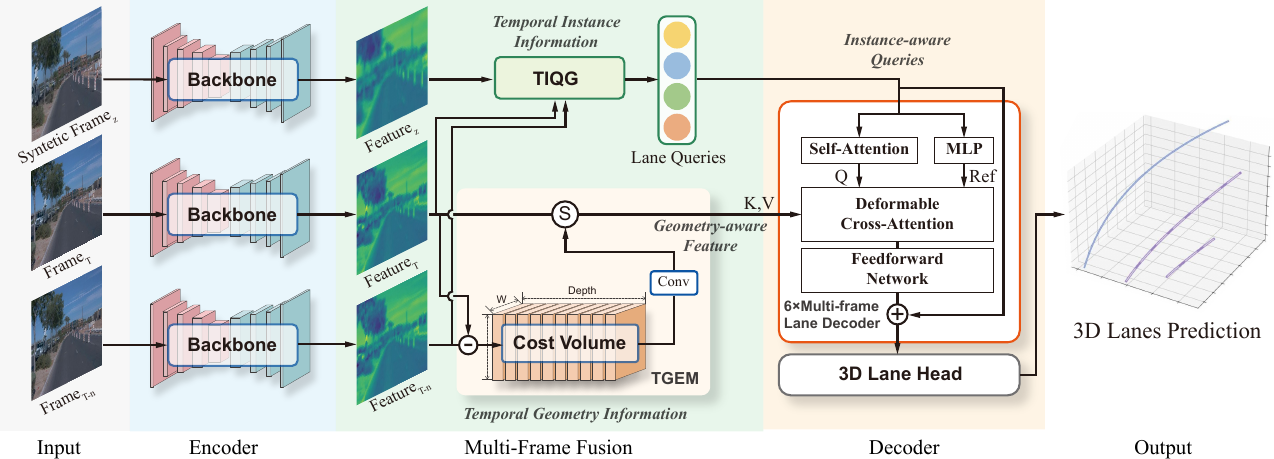}
    \vspace{-4mm}
    \caption{
        \textbf{Illustration of the Overall Architecture of the Proposed GTA-Net}. The input images are first fed into a backbone network to extract 2D perspective features. Next, TGEM is applied to enhance the geometric perception. Simultaneously, TIQG is used to generate lane queries. The lane queries and features are sent to the decoder to obtain 3D lane predictions.
    }
    \vspace{-4mm}
    \label{fig:arch}
\end{figure*}

\subsection{3D Lane Detection}
The objective of 3D lane detection is to identify lanes within 3D space and accurately estimate their coordinates \cite{ma2024monocular, zheng2024pvalane, li2023grouplane}.
In contrast to 2D lane detection, 3D lane detection methods need to perceive the geometric information of scenes.
Some methodologies try to enhance perception accuracy by leveraging both lidar and camera data \cite{luo2023dv, luo2022m}.
However, the acquisition and annotation expenses associated with multi-modal data are prohibitively high, constraining the practical implementation of such methodologies.
Hence, monocular 3D lane detection garners increased interest due to its provision of a cost-effective alternative to multi-modal methodologies \cite{ma2024monocular}.

Monocular 3D lane detection is challenging due to the absence of depth information.
To address this issue, SALAD \cite{yan2022once} incorporates a depth prediction branch to estimate depth, which is used to project 2D lanes to 3D space.
However, employing depth estimation in the process of detecting lanes may introduce cumulative errors and further exhibits inferior performance.
IPM-based approaches have become increasingly prominent owing to their advantageous geometric properties for lane depiction in BEV perspective \cite{ma2024monocular}.
3D-LaneNet \cite{garnett20193d} utilizes IPM for the conversion of FV features into BEV features, enabling bidirectional information flow between image-view and top-view representations.
Next,  these features are leveraged for learning the anchor offsets of lanes in the BEV space.
Gen-LaneNet \cite{guo2020gen} introduces a novel lane anchor representation guided by geometry in a newly defined coordinate system, which employs a specific geometric transformation to directly compute real 3D lane points from the network output. 
It is essential to emphasize that aligning the lane points with the underlying top-view features in the newly defined coordinate frame is crucial for establishing a generalized method capable of effectively handling diverse scenes.

To overcome IPM's reliance on the flat-ground assumption~\cite{chen2023efficient}, recent methods operate directly in 3D. 
Anchor3DLane~\cite{huang2023anchor3dlane}, for instance, forgoes the BEV representation by regressing 3D anchors directly, which are then projected onto the image to extract features.
LATR \cite{luo2023latr} uses a transformer-based architecture for monocular 3D lane detection.
Specifically, the authors design a lane-aware query generator that produces queries derived from 2D lane segmentation.
Besides, dynamic 3D ground PE generator is presented to embed 3D information during query updating.

\section{Method}
Our proposed Geometry-aware Temporal Aggregation Network (GTA-Net) leverages temporal evolution to address the core challenges of monocular 3D lane detection.
This section details its architecture and key components.
We first provide an overview of the entire GTA-Net framework.
Then, we elaborate on our two core contributions: the Temporal Geometry Enhancement Module (TGEM) for better geometry perception, and the Temporal Instance-aware Query Generation (TIQG) for enhancing lane integrity.
Finally, we describe the training objective and loss functions.

\subsection{Overall Architecture}
\label{sec:architecture}
Given successive front-view images $I_t$ and $I_{t-n} \in \mathbb{R}^{3 \times H \times W}$, the objective of monocular 3D lane detection is to estimate the 3D coordinates of the lane lines in the image $I_t$.
Each 3D lane can be described by a set of 3D points along with the corresponding lane category:
\begin{equation}
	L = (P, C), \quad P = {(x_i, y_i, z_i)}_{i=1}^N, \\
\end{equation}
where $L$ denotes a 3D lane, $C$ represents the corresponding lane category of $L$, $P$ is composed of $N$ 3D points with $N$ being a predefined parameter.
It is noted that the sequence $\{y_1, y^2, ..., y^N\}$ is often defined as a predetermined longitudinal coordinate \cite{garnett20193d,guo2020gen,chen2022persformer}.
Based on the above descriptions about lane representation, the 3D lanes in an input image $I_t$ can be represented as $G = \{L^1, L^2, ..., L^M\}$, where $G$ and $M$ represent the collection and the number of 3D lanes, respectively.

The overall architecture of our GTA-Net is illustrated in figure \ref{fig:arch}.
Initially, a 2D backbone network is utilized to extract image features $F_t$ and $F_{t-n}$ from the input front-view images $I_t$ and $I_{t-n}$:
\begin{equation}
	F_t = \text{Backbone}(I_t), \quad
	F_{t-n} = \text{Backbone}(I_{t-n}), \\
\end{equation}
where $\text{Backbone}(\cdot)$ denotes transformation of the used 2D backbone network.
Subsequently, the Temporal Geometry Enhancement Module (TGEM) takes these image features $F_t$ and $F_{t-n}$ as input to obtain geometry-aware features:
\begin{equation}
	F_{ge} = \text{TGEM}(F_{t-n}, F_t), \\
\end{equation}
where $F_{ge}$ and $\text{TGEM}(\cdot)$ indicate geometry-aware features and the transformation of TGEM, respectively.
TGEM aims to learn the geometric relationships between successive frames, thereby enhancing the depth perception capability of our GTA-Net in driving scenes.
Simultaneously, image features $F_t$ and $F_{t-n}$ are sent to Temporal Instance-aware Query Generation (TIQG), which generates queries by making full use of temporal instance information.
This process can be formulated as follows:
\begin{equation}
	Q = \text{TIQG}(F_{t-n}, F_t), \\
\end{equation}
where $Q$ represents the generated instance-aware queries, and $\text{TIQG}(\cdot)$ denotes the transformation performed by the TIQG module.
Next, the instance-aware queries $Q$ and geometry-aware features $F_{ge}$ are fed into a deformable attention \cite{xia2022vision, zhu2020deformable} based decoder to iteratively update queries:
\begin{equation}
	Q_u = \text{Decoder}(Q, F_{ge}), \\
\end{equation}
where $Q_u$ and $\text{Decoder}(\cdot)$ denote the updated queries and the transformation of the above deformable attention-based decoder, respectively.
Finally, the updated queries $Q_u$ are fed into the 3D lane detection head to output the predictions of 3D lanes in the input image $I_t$:
\begin{equation}
	\hat{G} = \text{Head}(Q_u), \\
\end{equation}
where $\hat{G}$ indicates the estimated 3D lanes, and $\text{Head}(\cdot)$ denotes 3D lane detection head. 
For the losses, we follow the setting of \cite{luo2023latr}.

\subsection{Temporal Geometry Enhancement Module}
\label{sec:TGEM}
The inherent absence of depth cues in monocular input images poses a significant challenge for monocular 3D lane detection solutions.
Without depth information, accurately understanding the complex geometric details of driving scenes becomes difficult. 
Drawing inspiration from prior research \cite{watson2021temporal, yang2020cost} in monocular depth estimation, this study aims to alleviate this limitation by utilizing the geometric relationship between consecutive frames.
To this end, we propose Temporal Geometry Enhancement Module (TGEM), which is designed to delve into the geometric consistency hidden in successive frames.
The structure of TGEM is described by figure \ref{fig:arch}.
In detail, TGEM consists of three main stages: cost volume construction, geometric feature leaning, geometry-aware feature enhancement.

\noindent\textbf{Cost volume construction.}
To uncover the potential geometric information within successive monocular images, we first construct a cost volume in the initial stage of TGEM. 
This is because the cost volume is capable of capturing the geometric relationship between adjacent frames \cite{gu2020cascade}.

Starting with 2D features $F_t$ and $F_{t-n}$ extracted from the 2D backbone network, we replicate these features $D$ times along the depth dimension, where $D$ represents a designated depth range.
The process can be formulated as follows:
\begin{equation}
	E_t = \text{Repeat}(F_t), \quad E_{t-n} = \text{Repeat}(F_{t-n}), \\
\end{equation}
where $E_t$ and $E_{t-n}$ denote the expanded features, and $\text{Repeat}(\cdot)$ represents the feature expanding operation.
Next, the expanded features $E_{t-n}$ in image coordinate system of $t-n$ frame is projected onto  image coordinate system of $t$ frame by using camera intrinsics, extrinsics and lidar poses:
\begin{equation}
	\hat{E_t} = \text{Warp}(F_{t-n}), \\
\end{equation}
where $\hat{E_t}$ denotes the warped features in image coordinate system of $t$ frame, and $\text{Warp}(\cdot)$ indicates the projection from image coordinate system of $t-n$ frame to image coordinate system of $t$ frame.
In the final step, the matching cost between $\hat{E_t}$ and $E_t$ is computed by using $l_1$ norm:
\begin{equation}
	C = \lVert \hat{E_t} - E_t \rVert_1, \\
\end{equation}
where ${\lVert \cdot \rVert}_1$ denotes the transformation of $l_1$ norm, and $C$ indicates the cost volume between the input successive front-view images $I_t$ and $I_{t-n}$.

\noindent\textbf{Geometric feature learning.}
Once the cost volume between successive front-view images is obtained, we utilize cascaded 2D convolution, batch normalization layer, and ReLU activation function to transform the cost volume into geometric features. 
This transformation can be expressed as follows:
\begin{equation}
	F_g = \text{ReLU}(\text{BN}(\text{Conv}(C))), \\
\end{equation}
where $F_g$ denotes geometric features containing the 3D structural information of the driving scene, and $\text{Conv}(\cdot)$, $\text{BN}(\cdot)$ and $\text{ReLU}(\cdot)$ represent the transformation of 2D convolution, batch normalization layer and ReLU activation function, respectively.

\begin{figure}[t]
	\centering
	\includegraphics[width=\linewidth]{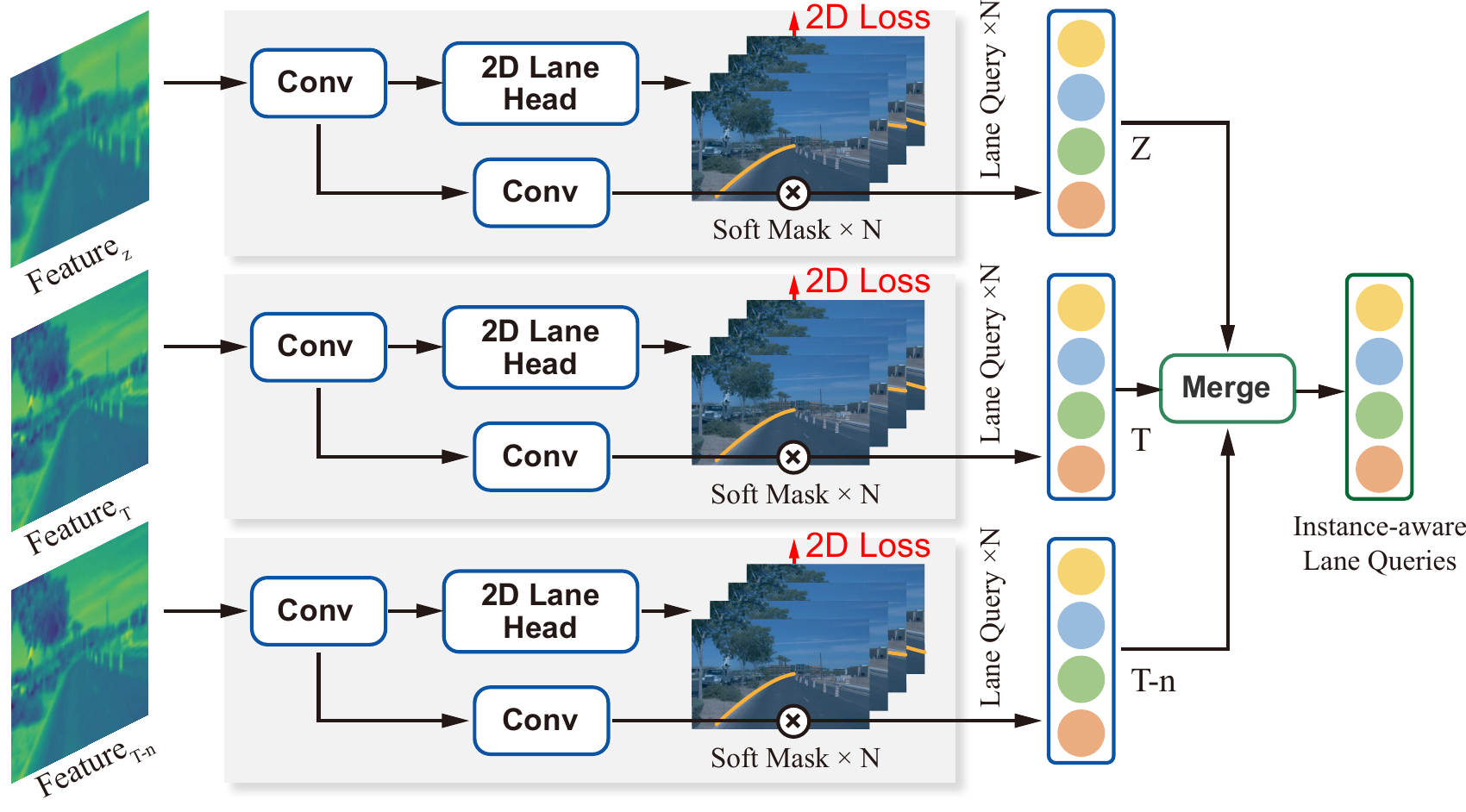} 
	\vspace{-6mm}
	\caption{\textbf{An Overview of the Proposed Temporal Instance-aware Query Generation (TIQG) Module.} TIQG strategically integrates temporal cues into the query generation process, thereby enhancing its capacity to exploit temporal instance information for promoting lane integrity.
	}
	\vspace{-5mm}
	\label{fig:tpig}  
\end{figure}

\noindent\textbf{Geometry-aware feature enhancement.}
To enrich the geometric perception capability of our GTA-Net, we aim to dynamically infuse geometric features into 2D features of $t$ frame.
To be specific, we first utilize geometric features $F_g$ to derive two learnable tensors: weights $\alpha$ and residual $\beta$:
\begin{equation}
	\alpha = \text{Sigmoid}(\text{BN}(\text{Conv}(F_g))), \\
\end{equation}
\begin{equation}
	\beta = \text{ReLU}(\text{BN}(\text{Conv}(F_g))), \\
\end{equation}
where $\text{Sigmoid}(\cdot)$ denotes the transformation of sigmoid activation function.
Subsequently, the weights $\alpha$ and residual $\beta$ are used to inject geometric information into 2D features:
\begin{equation}
	F_{ge} = \alpha F_t + \beta. \\
\end{equation}
As a result, the feature enhancement mechanism enables our GTA-Net to dynamically adjust and integrate geometric information into the feature extraction process, thereby enhancing its ability to perceive intricate geometric structures within the scene.

\subsection{Temporal Instance-aware Query Generation}
\label{sec:TIQG}
While the TGEM module provides rich geometric context (\textit{what the scene looks like}), it does not explicitly tell us \textit{where the lanes are}. 
Accurately localizing all lanes, especially those that are distant or partially occluded, requires instance-level priors.
Successive frames offer a wealth of such information: a lane barely visible now might have been clear seconds ago.
To harness this rich temporal context, we propose the Temporal Instance-aware Query Generation (TIQG) module, as illustrated in Figure~\ref{fig:tpig}.
The goal of TIQG is to generate a comprehensive and highly accurate set of lane queries by strategically aggregating information from the past, present, and a synthesized future.
This process unfolds in three key stages: initial query generation from monocular frames, synthetic future query generation for distant lanes, and finally, temporal query aggregation.

\noindent\textbf{Initial query generation.}
The initial query generation process involves generating temporal queries for both frame $t$ and frame $t-n$.
The queries consist of two-level embeddings: instance-level embeddings and point-level embeddings.
To derive instance-level embedding, cascaded 2D convolutions are employed to process image features $F_t$ and $F_{t-n}$ as follows:
\begin{equation}
	Q_t^{l} = \text{Conv}(F_t), \quad
	Q_{t-n}^{l} = \text{Conv}(F_{t-n}), \\
\end{equation}
where $Q_t^{l}$ and $Q_{t-n}^{l}$ denote the generated instance-level embeddings of frame $t$ and frame $t-n$, respectively.
For point-level embedding, we define two learnable parameters $Q_t^{p}$ and $Q_{t-n}^{p}$.
Subsequently, these corresponding two-level embeddings are merged using the following formulation:
\begin{equation}
	Q_t = Q_t^{l} \oplus Q_t^{p}, \quad
	Q_{t-n} = Q_{t-n}^{l} \oplus Q_{t-n}^{p},  \\
\end{equation}
where $\oplus$ represents the broadcasting summation.

\begin{figure}[t]
	\centering
	\includegraphics[width=\linewidth]{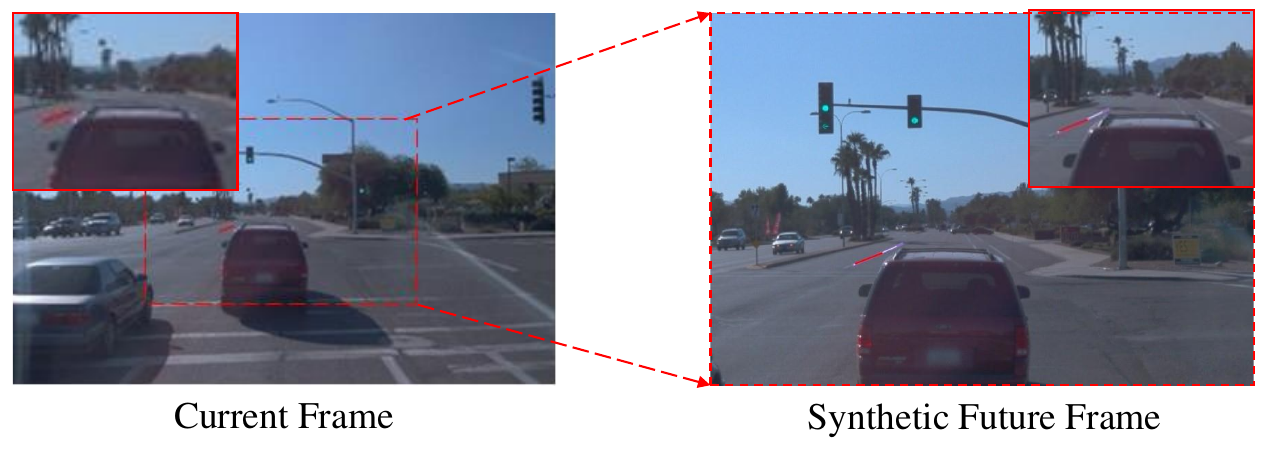}
	\vspace{-6mm}
	\caption{
		\textbf{Monocular Lane Detection Results of Setting Current and the Synthetic Future Frames as Input.}
		The white line represents the predicted lane line and the red line denotes the gt. We see that the lane line is missing in the predictions when accepting the current frame. In contrast, there is a white line in the visual results for treating the synthetic future frame as input.
	}
	\vspace{-5mm}
	\label{Fig:zoomin}
\end{figure}

\noindent\textbf{Synthetic future query generation.}
As shown in figure \ref{Fig:zoomin}, the 3D lane is not detected at the intersection, likely due to increased distance of the lane from the camera.
To mitigate this issue, we seek to reduce the distance between the camera and the lane.
A straightforward approach would involve using a future frame; however, real-world applications do not provide access to future information. 
As a solution, we synthesize a future frame by zooming in on the scene. 
This synthetic frame is then fed into the 3D lane detection model. 
Notably, we observe that lanes, which were initially undetected, become discernible in the synthetic future frame.
This observation motivates the generation of pseudo-future frame to effectively detect lanes distant regions.

Based on the aforementioned analysis, we design the synthetic future query generation.
Initially, we first crop and magnify the input image $I_t$ to generate a synthetic future frame.
Then, this synthetic future frame undergoes feature extraction to derive 2D perspective features $F_{p}$.
Once the 2D perspective features $F_{p}$ are obtained, the corresponding queries can be formulated as follows:
\begin{equation}
	Q_p^{l} = \text{Conv}(F_p), \\ \quad
	Q_p = Q_p^{l} \oplus Q_p^{p},
\end{equation}
where $Q_p$ denotes synthetic future queries, while $Q_p^{l}$ and $Q_p^{p}$ are instance-level embedding and point-level embedding, respectively.
By integrating both instance-level and point-level embeddings, TIQG gains a comprehensive understanding of the distant context, enhancing its capability to detect 3D lanes in distant regions.

\begin{table*}[t]
	\caption{
		\textbf{Quantitative Comparison of Monocular 3D Lane Detection Solutions on the OpenLane Dataset.} 
		$\downarrow$ and $\uparrow$ indicate that lower and higher values, respectively, signify superior performance. 
		\textbf{Bold} numbers represent the best results.
		GTA-Net obtains the competitive performance on F1-score, lane category accuracy, x/z errors for both near and far range.
	}
        \vspace{-2mm}
	\setlength{\tabcolsep}{16pt}
	\resizebox{\linewidth}{!}{
		\centering
		\begin{tabular}{lcccllll}
			\toprule[1.2pt]
			\multicolumn{1}{c}{} & \multicolumn{1}{c}{} & \multicolumn{1}{c}{} & {{Category } \multirow{-2}{*}{\quad}} & \multicolumn{2}{c}{{X error (m)} $\downarrow$} & \multicolumn{2}{c}{{Z error (m)} $\downarrow$} \\
			\cmidrule{4-5} \cmidrule{6-7} 
			\multicolumn{1}{l}{\multirow{-2}{*}{{Methods} \,}} & \multicolumn{1}{l}{\multirow{-2}{*}{{Year} \,}} & \multicolumn{1}{c}{\multirow{-2}{*}{{F1 (\%)} $\uparrow$}} & {{Accuracy (\%)} \multirow{-2}{*}{\, $\uparrow$}} & \multicolumn{1}{l}{\textit{near}} & \multicolumn{1}{l}{\,\textit{far}} & \multicolumn{1}{l}{\textit{near}} & \multicolumn{1}{l}{\,\textit{far}} \\ 
			\addlinespace[1mm]
			\hline \hline
			\addlinespace[1mm]
			3DLaneNet \cite{garnett20193d}   & 2019     & 44.1 &  -  \quad\quad\quad & 0.479 & 0.572 & 0.367 & 0.443 \\
			GenLaneNet \cite{guo2020gen}  & 2020     & 32.3 &  -  \quad\quad\quad & 0.593 & 0.494 & 0.140 & 0.195 \\
			Cond-IPM     & -     & 36.6 &  -  \quad\quad\quad & 0.563 & 1.080 & 0.421 & 0.892 \\
			Persformer \cite{chen2022persformer}   & 2022    & 50.5 & 89.5\quad\quad\quad & 0.319 & 0.325 & 0.112 & 0.141 \\
			CurveFormer \cite{bai2023curveformer}  & 2023     & 50.5 &  -  \quad\quad\quad & 0.340 & 0.772 & 0.207 & 0.651 \\
			BEV-LaneDet \cite{wang2022bev}  & 2022    & 58.4 &  -  \quad\quad\quad & 0.309 & 0.659 & 0.244 & 0.631 \\
			Anchor3DLane \cite{huang2023anchor3dlane}  & 2023   & 53.1 & 90.0\quad\quad\quad & 0.300 & 0.311 & 0.103 & 0.139 \\
			DecoupleLane \cite{han2023decoupling} & 2023 & 51.2 & -  \quad\quad\quad & - & - & - & - \\
			LaneCMK \cite{zhao2024lanecmkt} & 2024 & 55.8 & 89.2\quad\quad\quad  & 0.310 & 0.303 & 0.083 & 0.123 \\
			GroupLane \cite{li2024grouplane} & 2024 & 60.2 & 91.6\quad\quad\quad  & 0.371 & 0.476 & 0.220 & 0.357 \\
			CaliFree3DLane \cite{guo2024califree3dlane} & 2024 & 56.4 & -\quad\quad\quad & 0.319 & 0.767 & 0.245 & 0.706 \\
			CaliFree3DLane$^*$ \cite{guo2024califree3dlane} & 2024 & 57.0 &  -\quad\quad\quad & 0.306 & 0.761 & 0.227 & 0.683 \\
			3DLaneFormer \cite{dong20243dlaneformer} & 2024 & 55.0 & 89.9\quad\quad\quad & 0.283 & 0.305 & 0.106 & 0.138 \\
                DB3D-L \cite{liu2025db3d} & 2025 & 55.2 & - & 0.254 & 0.682 & 0.202 & 0.622 \\
                Freq-3DLane \cite{song2025freq} & 2025  & 59.7 & - & 0.265 & 0.665 & 0.196 & 0.608 \\
                CurveFormer++ \cite{bai2025curveformer++}  & 2025   & 52.5 & 87.8\quad\quad\quad & 0.333 & 0.805 & 0.186 & 0.687 \\
			\addlinespace[1mm]
			\hline \hline 
			\addlinespace[1mm]
			\rowcolor{gray!15}
			Ours         & -     & \textbf{62.4} & \textbf{92.8}\quad\quad\quad & \textbf{0.225} & \textbf{0.254} & \textbf{0.078} & \textbf{0.110} \\
			\bottomrule[1.2pt]
	\end{tabular}}
	\vspace{-1mm}
	\label{tab:openlane}
\end{table*}

\noindent\textbf{Temporal query aggregation.}
To fully explore temporal context, the information in these temporal queries $Q_t$, $Q_{t-1}$ and $Q_p$ are interacted by using cross-attention in the step of temporal query aggregation, which can be described by:
\begin{equation}
	Q_p^u = \text{CA}(Q_p, Q_t), \quad
	Q_{t-1}^u = \text{CA}(Q_{t-1}, Q_t), \\
\end{equation}
where $Q_p^u$, $Q_{t-1}^u$ and $\text{CA}(\cdot)$ denote the updated synthetic future queries, the updated queries of frame $t-1$ and the transformation of cross-attention.
Subsequently, these lane queries are fed into cascaded convolutions to obtain instance-aware queries $Q$:
\begin{equation}
	Q = \text{Conv}(\text{Concat}(Q_t, Q_{t-1}, Q_p)), \\
\end{equation}
where $\text{Concat}(\cdot)$ denotes concatenation operation. 
The instance-aware queries integrate rich instance-aware information, thereby enhancing the capacity to comprehensively perceive lanes to promote lane integrity.

\subsection{Loss Function}
\label{sec:loss}
With the matched ground truth 3D lanes corresponding to the output 3D lane queries, we compute the loss for each respective pair.
To be specific, L1 loss is employed to regress the $x$-coordinates and $z$-coordinates.
Binary cross-entropy loss is used to enforce the visibility constraint of the output 3D lanes.
Focal loss \cite{lin2017focal} is utilized to optimize the predicted lane category.
In addition, a 2D segmentation loss is also included as an auxiliary loss.
Hence, the final objective function of our GTA-Net can be formulated by:
\begin{align}
	\mathcal{L}_{total} = & \, \mathcal{L}_{vis} + w_{x}\mathcal{L}_x + w_{z}\mathcal{L}_z \notag \\
	& + w_{cate}\mathcal{L}_{cate} + w_{seg}\mathcal{L}_{seg}.
\end{align}
where $\mathcal{L}_{total}$ denotes the total loss, $w_x$, $w_z$, $w_{cate}$ and $w_{seg}$ are hyperparameters.

\begin{table*}[t]
	\caption{
		\textbf{Quantitative Comparison of Monocular 3D Lane Detection Methods on the OpenLane Dataset Under Different Scenarios.}
		\textbf{Bold} numbers denote the best results.
		"All" indicates the average value of F1 scores in all scenarios.
		The proposed GTA-Net achieves superior performance under various extremely challenging scenarios than other solutions.
	}
        \vspace{-2mm}
	\centering
	\setlength{\tabcolsep}{14pt}
	\resizebox{\linewidth}{!}{
		\begin{tabular}{l|c|c|c|c|c|c|c|c}
			\toprule[1.2pt]
			\multirow{2}{*}{Methods} & \multirow{2}{*}{Year} & \multirow{2}{*}{All} & Up \&  & \multirow{2}{*}{Curve} & Extreme & \multirow{2}{*}{Night} & \multirow{2}{*}{Intersection} & Merge  \\
			& & &  Down & & Weather &  & & \& Split\\
			\addlinespace[1mm]
			\hline \hline
			\addlinespace[1mm]
			3DLaneNet \cite{garnett20193d}   & 2019     & 44.1 & 40.8 & 46.5 & 47.5 & 41.5 & 32.1 & 41.7 \\
			GenLaneNet \cite{guo2020gen}    & 2020      & 32.3 & 25.4 & 33.5 & 28.1 & 18.7 & 21.4 & 31.0 \\
			Persformer \cite{chen2022persformer}    & 2022   & 50.5 & 42.4 & 55.6 & 48.6 & 46.6 & 40.0 & 50.7 \\
			CurveFormer \cite{bai2023curveformer}   & 2023    & 50.5 & 45.2 & 56.6 & 49.7 & 49.1 & 42.9 & 45.4 \\
			BEV-LaneDet \cite{wang2022bev}   & 2022   & 58.4 & 48.7 & 63.1 & 53.4 & 53.4 & 50.3 & 53.7 \\
			Anchor3DLane \cite{huang2023anchor3dlane}   & 2023  & 54.3 & 47.2 & 58.0 & 52.7 & 48.7 & 45.8 & 51.7 \\
			DecoupleLane \cite{han2023decoupling}  & 2023  & 51.2 & 43.5 & 57.3 &   -  & 48.9 & 43.5 &   -  \\
			LATR \cite{luo2023latr}      3& 2023       & 61.9 & 55.2 & 68.2 & 57.1 & 55.4 & 52.3 & 61.5 \\
			LaneCMK \cite{zhao2024lanecmkt} & 2024 & 55.8 & 47.3 & 58.6 & 53.2 & 48.0 & 42.2 & 51.7 \\
			CaliFree3DLane \cite{guo2024califree3dlane} & 2024 & 56.4 & 48.7 & 60.4 & 54.7 & 51.9 & 49.5 & 51.8 \\
			CaliFree3DLane$^*$ \cite{guo2024califree3dlane} & 2024 & 57.0& 48.9 & 62.3 & 54.8 & 52.0 & 50.2 & 52.1 \\
			3DLaneFormer \cite{dong20243dlaneformer} & 2024 & 55.0 & 49.0 & 58.3 & 54.5 & 50.4 & 46.3 & 52.3 \\
                Freq-3DLane \cite{song2025freq} & 2025 & 59.7 & 52.0 &  65.5 &  56.5 &  54.5 &  51.8 &  56.4 \\
			CurveFormer++ \cite{bai2025curveformer++} 5& 2025 & 52.7 & 48.3 & 59.4 & 50.6 & 48.4 & 45.0 & 48.1 \\
			\addlinespace[1mm]
			\hline \hline 
			\addlinespace[1mm]
			\rowcolor{gray!15}
			Ours & - & \textbf{62.4} & \textbf{55.5} & \textbf{68.3} & \textbf{57.5} & \textbf{56.8} & \textbf{52.7} & \textbf{61.8} \\
			\bottomrule[1.2pt]
		\end{tabular}
	}
	\label{tab:openlane-cases}
	\vspace{-3mm}
\end{table*}

\section{Experiments}
\label{sec: exp}
This section provides a detailed overview of our experimental setup.
We then evaluate the performance of our GTA-Net through a series of comparative analyses against state-of-the-art methods.
Additionally, we present several ablation studies to gain a deeper understanding of the effectiveness of our proposed approach.
Through these experiments, we aim to provide a thorough assessment of the capabilities and contributions of our model.

\subsection{Experimental Setup}

\textbf{Dataset and Evaluation metrics.}
In the experiments, we use the OpenLane dataset \cite{chen2022persformer}, derived from the Waymo dataset \cite{sun2020scalability}, as a comprehensive benchmark for 3D lane detection.
Comprising 1,000 segments and 200,000 frames, this dataset captures diverse environmental conditions such as varying weather, terrain, and lighting.
The images in the OpenLane dataset have a resolution of 1280$\times$1920 pixels and include 880,000 lane annotations spanning 14 categories.
The OpenLane dataset provides a diverse and challenging set for evaluating 3D lane detection algorithms.
We employ the official evaluation metrics to assess the effectiveness of our model on the OpenLane dataset.
Specifically, we report the F1 score, lane category accuracy, and x/z errors in both the near and far ranges.

\noindent\textbf{Implementation details.}
In our GTA-Net, we use ResNet-50 \cite{he2016deep} as the 2D backbone network for extracting image features.
Before feature extraction, all input images are resized to a resolution of 720$\times$960.
The experiments are conducted on the PyTorch platform using a workstation equipped with eight Tesla A100 GPUs, each with 40 GB of memory.
For optimization, we employ the AdamW optimizer with a weight decay of 0.01.
The initial learning rate is set to $2e^{-4}$, and a cosine annealing scheduler is applied to gradually adjust the learning rate. 
Our model is trained for 24 epochs, with a batch size of 64.
Additionally, we set $w_x=2$, $w_z=10$, $w_{cate}=10$ and $w_{seg}=5$.

\noindent\textbf{Compared methods.}
We compare our GTA-Net with the following monocular 3D detection methods: 
3DLaneNet \cite{garnett20193d}, Gen-LaneNet \cite{guo2020gen}, PersFormer \cite{chen2022persformer}, Cond-IPM, CurveFormer \cite{bai2023curveformer}, DecoupleLane \cite{han2023decoupling}, Anchor3DLane \cite{huang2023anchor3dlane}, BEV-LaneDet \cite{wang2022bev}, LATR \cite{luo2023latr}, GroupLane \cite{li2024grouplane}, LaneCMK \cite{zhao2024lanecmkt}, CaliFree3DLane$^*$ \cite{guo2024califree3dlane}, LaneCPP \cite{pittner2024lanecpp}, 3DLaneFormer \cite{dong20243dlaneformer}, DB3D-L \cite{liu2025db3d}, Freq-3DLane \cite{song2025freq} and CurveFormer++ \cite{bai2025curveformer++}.

\begin{figure*}[t]
	\centering
	\includegraphics[width=0.99\linewidth, height=0.26\linewidth]{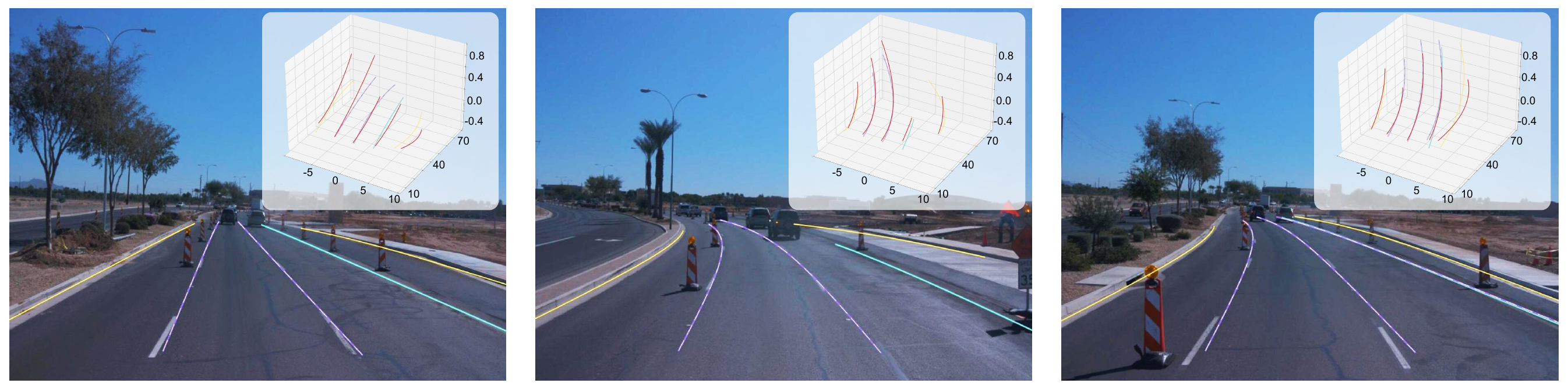}
	\vspace{-2mm}
	\caption{
		\textbf{Illustration of Predicted 3D Lanes on the OpenLane Dataset.}
		We present the predicted 3D lanes in both the perspective view and the 3D spatial representation. The red lines denote the ground truth lanes, while the other colored lines represent the predicted lanes.
	}
	\vspace{-3mm}
	\label{fig:openlane}
\end{figure*}

\subsection{Main Results}

\textbf{Quantitative results.}
We present a comprehensive quantitative evaluation of our proposed GTA-Net on the challenging OpenLane dataset, with detailed results summarized in Table \ref{tab:openlane}. The experimental findings unequivocally demonstrate that GTA-Net achieves superior performance, setting a new SoTA benchmark for monocular 3D lane detection.

Specifically, GTA-Net excels across key evaluation metrics. It attains the highest F1 score, indicative of its robust ability to precisely identify 3D lanes while effectively balancing recall and precision to minimize both missed detections and false positives. Furthermore, our method exhibits superior lane category accuracy, underscoring its enhanced capability in semantically understanding and classifying diverse lane types. Beyond detection, GTA-Net also demonstrates exceptional localization precision. It consistently yields the smallest $x$ and $z$ errors across both near and far perception ranges, a critical validation of its effectiveness in accurately estimating the spatial coordinates of 3D lanes, irrespective of their distance from the ego-vehicle. These results collectively affirm the robustness and practical utility of GTA-Net in complex scenarios.

To further validate the effectiveness of the proposed GTA-Net, we present results across a range of challenging scenarios. 
Table \ref{tab:openlane-cases} reports the 3D lane detection performance under different conditions, measured by the F1 score.
Notably, our method outperforms existing approaches, achieving the highest F1 score across all scenarios. 
This demonstrates the robustness of our model in handling diverse driving environments, underscoring its reliability and efficacy in accurately detecting 3D lanes under varying real-world conditions.

\noindent\textbf{Visual Results.} 
Beyond quantitative metrics, a rigorous qualitative assessment is crucial for a comprehensive understanding of GTA-Net's performance. 
To this end, we present a series of visualization results on the challenging OpenLane dataset, offering an intuitive insight into our model's capabilities. 
As vividly depicted in Figure \ref{fig:openlane}, GTA-Net consistently demonstrates an exceptional ability to accurately detect 3D lanes directly within the complex input images. 
More importantly, it precisely estimates their corresponding positions in 3D space, even under varying scene conditions. 
These visual outcomes not only corroborate the superior quantitative performance but also powerfully illustrate the model's robustness and accuracy in real-world scenarios, affirming its excellent capacity for both identifying and accurately localizing lanes in 3D space.

\begin{table}[t]
	\centering
        \small
	\caption{
		\textbf{Ablation study of the proposed modules.} We evaluate the effect of TGEM and TIQG by ablating them.
	}
	\vspace{-2mm}
	\setlength{\tabcolsep}{5pt}
    \renewcommand{\arraystretch}{1.3}
	\begin{tabular}{l c c c c c}
		\toprule[1.2pt]
		\multirow{2}{*}{Methods} & \multirow{2}{*}{F1 (\%)} & \multicolumn{2}{c}{X error (m)} & \multicolumn{2}{c}{Z error (m)} \\
		\cline{3-4} \cline{5-6}
		& & \textit{near} & \textit{far} & \textit{near} & \textit{far} \\
		\hline \hline
		Baseline                  & 61.7 & 0.235 & 0.273 & 0.085 & 0.120 \\
		\textit{w/o.} TGEM        & 62.1 & 0.229 & 0.257 & 0.081 & 0.112 \\
		\textit{w/o.} TIQG        & 62.1 & 0.229 & 0.264 & 0.080 & 0.114 \\
        \textit{w/o.} Cost Volume & 62.2 & 0.228 & 0.256 & 0.080 & 0.112 \\
		\hline \hline
		\rowcolor{gray!15}
		\textbf{Ours} & \textbf{62.4} & \textbf{0.225} & \textbf{0.254} & \textbf{0.078} & \textbf{0.110} \\
		\bottomrule[1.2pt]
	\end{tabular}
        \vspace{-4mm}
	\label{tab:ablation}
\end{table}

\subsection{Ablation Studies}
We design a series of meticulous experiments to evaluate the efficacy of our approach in addressing the inherent challenges of monocular 3D lane detection. 
These experiments are carefully structured to provide a comprehensive understanding of how the proposed techniques contribute to improving overall performance.

\noindent\textbf{Effect of TGEM.} To systematically dissect the contributions of individual components within our proposed GTA-Net framework, we first conduct an ablation study focusing on the Temporal Geometry Enhancement Module (TGEM). 
As detailed in Table \ref{tab:ablation}, we evaluate the performance of a modified GTA-Net variant, denoted as \textit{w/o.} TGEM, where the TGEM module is deliberately omitted. The comparative results reveal a marked and consistent degradation in performance across all evaluated metrics when TGEM is excluded. 
This significant decline unequivocally highlights the critical and indispensable role of TGEM. 
It actively promotes the network's ability to accurately capture and robustly leverage intricate geometric cues from the temporal evolution of the scene, which are crucial for resolving ambiguities inherent in monocular 3D lane detection.

\noindent\textbf{Effect of TIQG.}
We next evaluate the contribution of the TIQG module.
As presented in Table \ref{tab:ablation}, the variant labeled \textit{w/o.} TIQG corresponds to a version of GTA-Net where the TIQG module is omitted.
The results reveal a significant degradation in performance across all metrics when TIQG is excluded, thereby suggesting its critical role in effectively leveraging temporal cues. 
This demonstrates the importance of TIQG in improving the accuracy of 3D lane detection from monocular inputs by harnessing temporal instance information to enhance the integrity of lanes.

\paragraph{Effect of Cost Volume.}
To investigate the contribution of the cost volume construction within TGEM, we conduct an ablation study where this component is omitted. 
In this variant, referred to as \textit{w/o.} Cost Volume, the geometric feature learning directly processes the concatenated raw 2D features without an explicit cost volume generation step. 
As shown in Table \ref{tab:ablation}, the removal of the cost volume leads to a degradation in performance across all metrics. 
This suggests that while cost volume plays a role in capturing precise geometric relationships between frames. 
The cost volume provides a fine-grained inter-frame geometric correspondence that aids in depth-aware feature generation, and its absence impacts the overall accuracy.

\section{Conclusion}
In this paper, we tackle two key challenges in monocular 3D lane detection: the inaccurate geometric information of predicted 3D lanes and the difficulty in detecting remote lanes. 
To address these issues, we propose a novel Geometry-aware Temporal Aggregation Network (GTA-Net) by harnessing temporal evolution for monocular 3D lane detection.
We introduce the Temporal Geometry Enhancement Module (TGEM), which leverages geometric consistency across successive frames to enhance depth perception and improve geometric understanding of the scene. 
Additionally, we present the Temporal Instance-aware Query Generation (TIQG) module, which strategically incorporates temporal context into the query generation process, enabling a more comprehensive exploration of instance-aware information across frames.
Through extensive and rigorous experimentation, the results demonstrate that the proposed GTA-Net outperforms existing methods.

{
    \small
    \bibliographystyle{ieeenat_fullname}
    \bibliography{main}

@inproceedings{yan2022once,
  title={Once-3dlanes: Building monocular 3d lane detection},
  author={Yan, Fan and Nie, Ming and Cai, Xinyue and Han, Jianhua and Xu, Hang and Yang, Zhen and Ye, Chaoqiang and Fu, Yanwei and Mi, Michael Bi and Zhang, Li},
  booktitle={Proceedings of the IEEE/CVF Conference on Computer Vision and Pattern Recognition},
  year={2022}
}

@article{li2024grouplane,
	title={Grouplane: End-to-end 3d lane detection with channel-wise grouping},
	author={Li, Zhuoling and Han, Chunrui and Ge, Zheng and Yang, Jinrong and Yu, En and Wang, Haoqian and Zhang, Xiangyu and Zhao, Hengshuang},
	journal={IEEE Robotics and Automation Letters},
	year={2024},
}

@inproceedings{dong20243dlaneformer,
	title={3Dlaneformer: Rethinking Learning Views for 3D Lane Detection},
	author={Dong, Kun and Xue, Jian and Lan, Xing and Lu, Ke},
	booktitle={IEEE International Conference on Image Processing},
	year={2024},
}

@inproceedings{yan2025drivingsphere,
  title={Drivingsphere: Building a high-fidelity 4d world for closed-loop simulation},
  author={Yan, Tianyi and Wu, Dongming and Han, Wencheng and Jiang, Junpeng and Zhou, Xia and Zhan, Kun and Xu, Cheng-zhong and Shen, Jianbing},
  booktitle={Proceedings of the Computer Vision and Pattern Recognition Conference},
  year={2025}
}

@inproceedings{yan2025olidm,
  title={OLiDM: Object-aware LiDAR Diffusion Models for Autonomous Driving},
  author={Yan, Tianyi and Yin, Junbo and Lang, Xianpeng and Yang, Ruigang and Xu, Cheng-Zhong and Shen, Jianbing},
  booktitle={Proceedings of the AAAI Conference on Artificial Intelligence},
  year={2025}
}

@article{yan2025rlgf,
  title={RLGF: Reinforcement Learning with Geometric Feedback for Autonomous Driving Video Generation},
  author={Yan, Tianyi and Han, Wencheng and Zhou, Xia and Zhang, Xueyang and Zhan, Kun and Xu, Cheng-zhong and Shen, Jianbing},
  journal={arXiv preprint arXiv:2509.16500},
  year={2025}
}

@article{guo2024califree3dlane,
	title={CaliFree3DLane: Calibration Free Spatio-Temporal BEV Representation for Monocular 3D Lane Detection},
	author={Guo, Weizhi and Li, Chaochao and Li, Kaijiang and Lv, Pei and Xu, Mingliang},
	journal={IEEE Transactions on Intelligent Transportation Systems},
	year={2024},
}

@inproceedings{pittner2024lanecpp,
	title={LaneCPP: Continuous 3D Lane Detection using Physical Priors},
	author={Pittner, Maximilian and Janai, Joel and Condurache, Alexandru P},
	booktitle={Proceedings of the IEEE/CVF Conference on Computer Vision and Pattern Recognition},
	year={2024}
}

@article{song2025freq,
  title={Freq-3DLane: 3D Lane Detection From Monocular Images via Frequency-Aware Feature Fusion},
  author={Song, Yongchao and Bi, Jiping and Sun, Lijun and Liu, Zhaowei and Jiang, Yahong and Wang, Xuan},
  journal={IEEE Transactions on Intelligent Transportation Systems},
  year={2025},
}

@article{liu2025db3d,
  title={DB3D-L: Depth-aware BEV Feature Transformation for Accurate 3D Lane Detection},
  author={Liu, Yehao and Xu, Xiaosu and Wang, Zijian and Yao, Yiqing},
  journal={arXiv preprint arXiv:2505.13266},
  year={2025}
}

@inproceedings{zhao2024lanecmkt,
	title={LaneCMKT: Boosting Monocular 3D Lane Detection with Cross-Modal Knowledge Transfer},
	author={Zhao, Runkai and Wang, Heng and Cai, Weidong},
	booktitle={Proceedings of the 32nd ACM International Conference on Multimedia},
	year={2024}
}

@article{ma2024monocular,
  title={Monocular 3D lane detection for Autonomous Driving: Recent Achievements, Challenges, and Outlooks},
  author={Ma, Fulong and Qi, Weiqing and Zhao, Guoyang and Zheng, Linwei and Wang, Sheng and Liu, Ming},
  journal={arXiv preprint arXiv:2404.06860},
  year={2024}
}

@article{williams2022trajectory,
  title={Trajectory planning with deep reinforcement learning in high-level action spaces},
  author={Williams, Kyle R and Schlossman, Rachel and Whitten, Daniel and Ingram, Joe and Musuvathy, Srideep and Pagan, James and Williams, Kyle A and Green, Sam and Patel, Anirudh and Mazumdar, Anirban and others},
  journal={IEEE Transactions on Aerospace and Electronic Systems},
  year={2022},
}

@inproceedings{li2022hdmapnet,
  title={Hdmapnet: An online hd map construction and evaluation framework},
  author={Li, Qi and Wang, Yue and Wang, Yilun and Zhao, Hang},
  booktitle={International Conference on Robotics and Automation},
  year={2022},
}

@article{wang2005lane,
  title={Lane keeping based on location technology},
  author={Wang, Jin and Schroedl, Stefan and Mezger, Klaus and Ortloff, Roland and Joos, Armin and Passegger, Thomas},
  journal={IEEE Transactions on Intelligent Transportation Systems},
  year={2005},
}

@inproceedings{liu2023vectormapnet,
  title={Vectormapnet: End-to-end vectorized hd map learning},
  author={Liu, Yicheng and Yuan, Tianyuan and Wang, Yue and Wang, Yilun and Zhao, Hang},
  booktitle={International Conference on Machine Learning},
  year={2023},
}

@inproceedings{chen2022persformer,
  title={Persformer: 3d lane detection via perspective transformer and the openlane benchmark},
  author={Chen, Li and Sima, Chonghao and Li, Yang and Zheng, Zehan and Xu, Jiajie and Geng, Xiangwei and Li, Hongyang and He, Conghui and Shi, Jianping and Qiao, Yu and others},
  booktitle={European Conference on Computer Vision},
  year={2022},
}

@inproceedings{garnett20193d,
  title={3d-lanenet: end-to-end 3d multiple lane detection},
  author={Garnett, Noa and Cohen, Rafi and Pe'er, Tomer and Lahav, Roee and Levi, Dan},
  booktitle={Proceedings of the IEEE/CVF International Conference on Computer Vision},
  year={2019}
}

@inproceedings{guo2020gen,
  title={Gen-lanenet: A generalized and scalable approach for 3d lane detection},
  author={Guo, Yuliang and Chen, Guang and Zhao, Peitao and Zhang, Weide and Miao, Jinghao and Wang, Jingao and Choe, Tae Eun},
  booktitle={European Conference on Computer Vision},
  year={2020},
}

@inproceedings{wang2022bev,
  title={BEV-LaneDet: a Simple and Effective 3D Lane Detection Baseline},
  author={Wang, Ruihao and Qin, Jian and Li, Kaiying and Li, Yaochen and Cao, Dong and Xu, Jintao},
  booktitle={Proceedings of the IEEE/CVF Conference on Computer Vision and Pattern Recognition},
  year={2023}
}

@inproceedings{huang2023anchor3dlane,
  title={Anchor3dlane: Learning to regress 3d anchors for monocular 3d lane detection},
  author={Huang, Shaofei and Shen, Zhenwei and Huang, Zehao and Ding, Zi-han and Dai, Jiao and Han, Jizhong and Wang, Naiyan and Liu, Si},
  booktitle={Proceedings of the IEEE/CVF Conference on Computer Vision and Pattern Recognition},
  year={2023}
}

@inproceedings{luo2023latr,
  title={Latr: 3d lane detection from monocular images with transformer},
  author={Luo, Yueru and Zheng, Chaoda and Yan, Xu and Kun, Tang and Zheng, Chao and Cui, Shuguang and Li, Zhen},
  booktitle={Proceedings of the IEEE/CVF International Conference on Computer Vision},
  year={2023}
}

@article{bai2025curveformer++,
  title={Curveformer++: 3d lane detection by curve propagation with temporal curve queries and attention},
  author={Bai, Yifeng and Chen, Zhirong and Liang, Pengpeng and Song, Bo and Cheng, Erkang},
  journal={IEEE Transactions on Intelligent Transportation Systems},
  year={2025},
  publisher={IEEE}
}

@inproceedings{bai2023curveformer,
  title={Curveformer: 3d lane detection by curve propagation with curve queries and attention},
  author={Bai, Yifeng and Chen, Zhirong and Fu, Zhangjie and Peng, Lang and Liang, Pengpeng and Cheng, Erkang},
  booktitle={IEEE International Conference on Robotics and Automation},
  year={2023},
}

@article{tang2021review,
  title={A review of lane detection methods based on deep learning},
  author={Tang, Jigang and Li, Songbin and Liu, Peng},
  journal={Pattern Recognition},
  year={2021},
}

@article{ko2021key,
  title={Key points estimation and point instance segmentation approach for lane detection},
  author={Ko, Yeongmin and Lee, Younkwan and Azam, Shoaib and Munir, Farzeen and Jeon, Moongu and Pedrycz, Witold},
  journal={IEEE Transactions on Intelligent Transportation Systems},
  year={2021},
}

@article{li2021abssnet,
  title={ABSSNet: Attention-based spatial segmentation network for traffic scene understanding},
  author={Li, Xuelong and Zhao, Zhiyuan and Wang, Qi},
  journal={IEEE Transactions on Cybernetics},
  year={2021},
}

@inproceedings{zheng2021resa,
  title={Resa: Recurrent feature-shift aggregator for lane detection},
  author={Zheng, Tu and Fang, Hao and Zhang, Yi and Tang, Wenjian and Yang, Zheng and Liu, Haifeng and Cai, Deng},
  booktitle={Proceedings of the AAAI Conference on Artificial Intelligence},
  year={2021}
}

@inproceedings{jin2022eigenlanes,
  title={Eigenlanes: Data-driven lane descriptors for structurally diverse lanes},
  author={Jin, Dongkwon and Park, Wonhui and Jeong, Seong-Gyun and Kwon, Heeyeon and Kim, Chang-Su},
  booktitle={Proceedings of the IEEE/CVF Conference on Computer Vision and Pattern Recognition},
  year={2022}
}

@inproceedings{liu2021condlanenet,
  title={Condlanenet: a top-to-down lane detection framework based on conditional convolution},
  author={Liu, Lizhe and Chen, Xiaohao and Zhu, Siyu and Tan, Ping},
  booktitle={Proceedings of the IEEE/CVF international conference on computer vision},
  year={2021}
}

@inproceedings{qin2020ultra,
  title={Ultra fast structure-aware deep lane detection},
  author={Qin, Zequn and Wang, Huanyu and Li, Xi},
  booktitle={European Conference on Computer Vision},
  year={2020},
}

@article{qin2022ultra,
  title={Ultra fast deep lane detection with hybrid anchor driven ordinal classification},
  author={Qin, Zequn and Zhang, Pengyi and Li, Xi},
  journal={IEEE Transactions on Pattern Analysis and Machine Intelligence},
  year={2022},
}

@inproceedings{tabelini2021keep,
  title={Keep your eyes on the lane: Real-time attention-guided lane detection},
  author={Tabelini, Lucas and Berriel, Rodrigo and Paixao, Thiago M and Badue, Claudine and De Souza, Alberto F and Oliveira-Santos, Thiago},
  booktitle={Proceedings of the IEEE/CVF Conference on Computer Vision and Pattern Recognition},
  year={2021}
}

@inproceedings{yoo2020end,
  title={End-to-end lane marker detection via row-wise classification},
  author={Yoo, Seungwoo and Lee, Hee Seok and Myeong, Heesoo and Yun, Sungrack and Park, Hyoungwoo and Cho, Janghoon and Kim, Duck Hoon},
  booktitle={Proceedings of the IEEE/CVF Conference on Computer Vision and Pattern Recognition workshops},
  year={2020}
}

@inproceedings{qu2021focus,
  title={Focus on local: Detecting lane marker from bottom up via key point},
  author={Qu, Zhan and Jin, Huan and Zhou, Yang and Yang, Zhen and Zhang, Wei},
  booktitle={Proceedings of the IEEE/CVF Conference on Computer Vision and Pattern Recognition},
  year={2021}
}

@inproceedings{wang2022keypoint,
  title={A keypoint-based global association network for lane detection},
  author={Wang, Jinsheng and Ma, Yinchao and Huang, Shaofei and Hui, Tianrui and Wang, Fei and Qian, Chen and Zhang, Tianzhu},
  booktitle={Proceedings of the IEEE/CVF Conference on Computer Vision and Pattern Recognition},
  year={2022}
}

@inproceedings{feng2022rethinking,
  title={Rethinking efficient lane detection via curve modeling},
  author={Feng, Zhengyang and Guo, Shaohua and Tan, Xin and Xu, Ke and Wang, Min and Ma, Lizhuang},
  booktitle={Proceedings of the IEEE/CVF Conference on Computer Vision and Pattern Recognition},
  year={2022}
}

@inproceedings{liu2021end,
  title={End-to-end lane shape prediction with transformers},
  author={Liu, Ruijin and Yuan, Zejian and Liu, Tie and Xiong, Zhiliang},
  booktitle={Proceedings of the IEEE/CVF winter conference on applications of computer vision},
  year={2021}
}

@inproceedings{tabelini2021polylanenet,
  title={Polylanenet: Lane estimation via deep polynomial regression},
  author={Tabelini, Lucas and Berriel, Rodrigo and Paixao, Thiago M and Badue, Claudine and De Souza, Alberto F and Oliveira-Santos, Thiago},
  booktitle={International Conference on Pattern Recognition},
  year={2021},
}

@inproceedings{luo2023dv,
  title={DV-3DLane: End-to-end Multi-modal 3D Lane Detection with Dual-view Representation},
  author={Luo, Yueru and Cui, Shuguang and Li, Zhen},
  booktitle={International Conference on Learning Representations},
  year={2023}
}

@article{luo2022m,
  title={Mˆ2-3dlanenet: Multi-modal 3d lane detection},
  author={Luo, Yueru and Yan, Xu and Zheng, Chaoda and Zheng, Chao and Mei, Shuqi and Kun, Tang and Cui, Shuguang and Mˆ, Zhen Li},
  journal={arXiv preprint arXiv:2209.05996},
  year={2022}
}

@inproceedings{xia2022vision,
  title={Vision transformer with deformable attention},
  author={Xia, Zhuofan and Pan, Xuran and Song, Shiji and Li, Li Erran and Huang, Gao},
  booktitle={Proceedings of the IEEE/CVF Conference on Computer Vision and Pattern Recognition},
  year={2022}
}

@article{zhu2020deformable,
  title={Deformable detr: Deformable transformers for end-to-end object detection},
  author={Zhu, Xizhou and Su, Weijie and Lu, Lewei and Li, Bin and Wang, Xiaogang and Dai, Jifeng},
  journal={arXiv preprint arXiv:2010.04159},
  year={2020}
}

@inproceedings{watson2021temporal,
  title={The temporal opportunist: Self-supervised multi-frame monocular depth},
  author={Watson, Jamie and Mac Aodha, Oisin and Prisacariu, Victor and Brostow, Gabriel and Firman, Michael},
  booktitle={Proceedings of the IEEE/CVF Conference on Computer Vision and Pattern Recognition},
  year={2021}
}

@inproceedings{yang2020cost,
  title={Cost volume pyramid based depth inference for multi-view stereo},
  author={Yang, Jiayu and Mao, Wei and Alvarez, Jose M and Liu, Miaomiao},
  booktitle={Proceedings of the IEEE/CVF Conference on Computer Vision and Pattern Recognition},
  year={2020}
}

@inproceedings{gu2020cascade,
  title={Cascade cost volume for high-resolution multi-view stereo and stereo matching},
  author={Gu, Xiaodong and Fan, Zhiwen and Zhu, Siyu and Dai, Zuozhuo and Tan, Feitong and Tan, Ping},
  booktitle={Proceedings of the IEEE/CVF Conference on Computer Vision and Pattern Recognition},
  year={2020}
}

@inproceedings{sun2020scalability,
  title={Scalability in perception for autonomous driving: Waymo open dataset},
  author={Sun, Pei and Kretzschmar, Henrik and Dotiwalla, Xerxes and Chouard, Aurelien and Patnaik, Vijaysai and Tsui, Paul and Guo, James and Zhou, Yin and Chai, Yuning and Caine, Benjamin and others},
  booktitle={Proceedings of the IEEE/CVF Conference on Computer Vision and Pattern Recognition},
  year={2020}
}

@inproceedings{he2016deep,
  title={Deep residual learning for image recognition},
  author={He, Kaiming and Zhang, Xiangyu and Ren, Shaoqing and Sun, Jian},
  booktitle={Proceedings of the IEEE conference on computer vision and pattern recognition},
  year={2016}
}

@article{han2023decoupling,
  title={Decoupling the Curve Modeling and Pavement Regression for Lane Detection},
  author={Han, Wencheng and Shen, Jianbing},
  journal={arXiv preprint arXiv:2309.10533},
  year={2023}
}

@inproceedings{lin2017focal,
  title={Focal loss for dense object detection},
  author={Lin, Tsung-Yi and Goyal, Priya and Girshick, Ross and He, Kaiming and Doll{\'a}r, Piotr},
  booktitle={Proceedings of the IEEE international conference on computer vision},
  year={2017}
}

@inproceedings{zheng2024pvalane,
  title={PVALane: Prior-Guided 3D Lane Detection with View-Agnostic Feature Alignment},
  author={Zheng, Zewen and Zhang, Xuemin and Mou, Yongqiang and Gao, Xiang and Li, Chengxin and Huang, Guoheng and Pun, Chi-Man and Yuan, Xiaochen},
  booktitle={Proceedings of the AAAI Conference on Artificial Intelligence},
  year={2024}
}

@article{li2023grouplane,
  title={Grouplane: End-to-end 3d lane detection with channel-wise grouping},
  author={Li, Zhuoling and Han, Chunrui and Ge, Zheng and Yang, Jinrong and Yu, En and Wang, Haoqian and Zhang, Xiangyu and Zhao, Hengshuang},
  journal={IEEE Robotics and Automation Letters},
  year={2024},
}

@article{chen2023efficient,
  title={An efficient transformer for simultaneous learning of BEV and lane representations in 3D lane detection},
  author={Chen, Ziye and Smith-Miles, Kate and Du, Bo and Qian, Guoqi and Gong, Mingming},
  journal={arXiv preprint arXiv:2306.04927},
  year={2023}
}

@inproceedings{jin2023recursive,
  title={Recursive Video Lane Detection},
  author={Jin, Dongkwon and Kim, Dahyun and Kim, Chang-Su},
  booktitle={Proceedings of the IEEE/CVF International Conference on Computer Vision},
  year={2023}
}

@inproceedings{zheng2022clrnet,
  title={Clrnet: Cross layer refinement network for lane detection},
  author={Zheng, Tu and Huang, Yifei and Liu, Yang and Tang, Wenjian and Yang, Zheng and Cai, Deng and He, Xiaofei},
  booktitle={Proceedings of the IEEE/CVF Conference on Computer Vision and Pattern Recognition},
  year={2022}
}

@inproceedings{han2022laneformer,
  title={Laneformer: Object-aware row-column transformers for lane detection},
  author={Han, Jianhua and Deng, Xiajun and Cai, Xinyue and Yang, Zhen and Xu, Hang and Xu, Chunjing and Liang, Xiaodan},
  booktitle={Proceedings of the AAAI conference on artificial intelligence},
  year={2022}
}

@inproceedings{xu2020curvelane,
  title={Curvelane-nas: Unifying lane-sensitive architecture search and adaptive point blending},
  author={Xu, Hang and Wang, Shaoju and Cai, Xinyue and Zhang, Wei and Liang, Xiaodan and Li, Zhenguo},
  booktitle={Computer Vision--ECCV 2020: 16th European Conference, Glasgow, UK, August 23--28, 2020, Proceedings, Part XV 16},
  year={2020},
}

@inproceedings{kirillov2023segment,
  title={Segment anything},
  author={Kirillov, Alexander and Mintun, Eric and Ravi, Nikhila and Mao, Hanzi and Rolland, Chloe and Gustafson, Laura and Xiao, Tete and Whitehead, Spencer and Berg, Alexander C and Lo, Wan-Yen and others},
  booktitle={Proceedings of the IEEE/CVF International Conference on Computer Vision},
  year={2023}
}
}


\end{document}